\pdfoutput=1
\documentclass[11pt]{article}

\usepackage[final]{acl}
\usepackage{times}
\usepackage{latexsym}
\usepackage{subcaption}
\usepackage[T1]{fontenc}
\usepackage[utf8]{inputenc}
\usepackage{microtype}
\usepackage{inconsolata}
\usepackage{graphicx}
\usepackage{listings}
\usepackage{xcolor}
\usepackage{multirow}
\usepackage{threeparttable}
\usepackage{ragged2e}
\usepackage{pifont}
\newcommand{\cmark}{\ding{51}}

\definecolor{blue}{HTML}{0068A9}
\definecolor{red}{HTML}{C23538}
\definecolor{green}{HTML}{3D8A2F}
\definecolor{orange}{HTML}{E96301}
\definecolor{purple}{HTML}{792C74}

\title{\textbf{PDFMathTranslate: Scientific Document Translation Preserving Layouts}}

\author{
    Rongxin Ouyang \\
      National University of Singapore / Blk AS6, \#03-41, 11 Computing Drive \\
      Singapore 117416, Singapore \\
      \texttt{rongxin@u.nus.edu}
    \AND
      Chang Chu\\
      Tsinghua University, Shenzhen International Graduate School\\
      Shenzhen China
      \texttt{chuc23@mails.tsinghua.edu.cn}
     \AND
    Zhikuang Xin \\
      University of Chinese Academy of Sciences \\ Computer Network Information Center, Chinese Academy of Sciences, \\
      Beijing China\\
      \texttt{xinzhikuang@cnic.cn} \\
    \AND
    Xiangyao Ma \\
      Funstory.ai Limited / HK \\
      \texttt{aw@funstory.ai} \\
}
  
\author{
    \textbf{Rongxin Ouyang\textsuperscript{1,$\dagger$}},
    \textbf{Chang Chu\textsuperscript{2,$\dagger$,$\ast$}},
    \textbf{Zhikuang Xin\textsuperscript{3}},
    \textbf{Xiangyao Ma\textsuperscript{4}}
    \\
    \textsuperscript{1}National University of Singapore, Singapore.
    \textsuperscript{2}Tsinghua University, Beijing, China.\\
    \textsuperscript{3}University of Chinese Academy of Sciences, Beijing, China.\\
    \textsuperscript{4}Funstory.ai Limited, Hong Kong SAR, China.\\
    \textsuperscript{$\dagger$} These authors contribute equally to this work.\\
    \textsuperscript{$\ast$}\textbf{Correspondence:} \href{mailto:chuc23@mails.tsinghua.edu.cn}{chuc23@mails.tsinghua.edu.cn}.
}

\begin{document}
\maketitle
\begin{abstract}

Language barriers in scientific documents hinder the diffusion and development of science and technologies. However, prior efforts in translating such documents largely overlooked the information in layouts. To bridge the gap, we introduce \textbf{PDFMathTranslate}, the world's first open-source software for translating scientific documents while preserving layouts. Leveraging the most recent advances in large language models and precise layout detection, we contribute to the community with key improvements in precision, flexibility, and efficiency. The work is open-sourced at~\href{https://github.com/byaidu/pdfmathtranslate}{https://github.com/byaidu/pdfmathtranslate} with more than 222k downloads.
% (1) efficient workflow; (2) support for multiple languages; (3) support for multiple models and services; (4) diverse interfaces; and (5) a sustainable development model. 
\end{abstract}

\section{Introduction}

Science and technologies diffuse and develop in different languages ~\cite{vongizycki1973centre, montgomery2013does}. Yet, language barriers hinder the diffusion and development of scientific progress~\cite{ramirez2020disadvantages, hwang2005inferior, ulrichammon2012linguistic}. For example, while 98\% publications in science are written in English~\cite{liu2017changing, ulrichammon2012linguistic}, the language only has 7.3\% native speakers and no more than 20\% speakers worldwide~\cite{bahji2023exclusion}. Consequently, the majority of human beings are hindered by language barriers from the advances in science and technologies. To overcome the barrier, there are the attempts from international organizations for inclusive science \footnote{See \url{https://www.unesco.org/en/open-science}} and the attempts from academia in improving \textit{Machine Translation (MT)} over text~\cite{brown1990statistical, vaswani2017attention,zhu2020incorporating, johnson2017googles, sennrich2016neural}.

However, text-based machine translation fails to address the unique challenges posed by the layouts in \textit{technical translation}~\cite{schubert2012technical} where the \textit{elaborate access structure} matters (pp. 351-352). Non-textual elements in scientific and technical documents are not ignorable --- the arrangement of paragraphs, mathematical equations, tables, and figures have rich and important meanings. Thus, text-based machine translation ignoring such important information appear insufficient to address the barrier in scientific and technical document translation, hindering the progress of science, technologies, and society.

% Attempts to address the format challenge are limited in either quality or scope. The first line of research 

% image-based translation [citation add] considering visual representations, these models, however, are limited in following the improving quality of large language models (LLM) \cite{dalayli-2023-use}. On the other hand, commercial services are not flexible and economical enough in many scenarios, as suggested in our comparison. Moreover, to our knowledge, there are no mature open-source solutions - that is, both veins of solutions are limited to a tiny group of human beings and are insufficient to address the language issue existing in the majority of the human population.
% \textit{Portable Document Format (PDF)} is one of the most common file types for the publication and distribution of scientific documents [add cite], with complex features in formatting [add cite],

\begin{figure*}[htp!]
    \centering
    \includegraphics[width=1\linewidth]{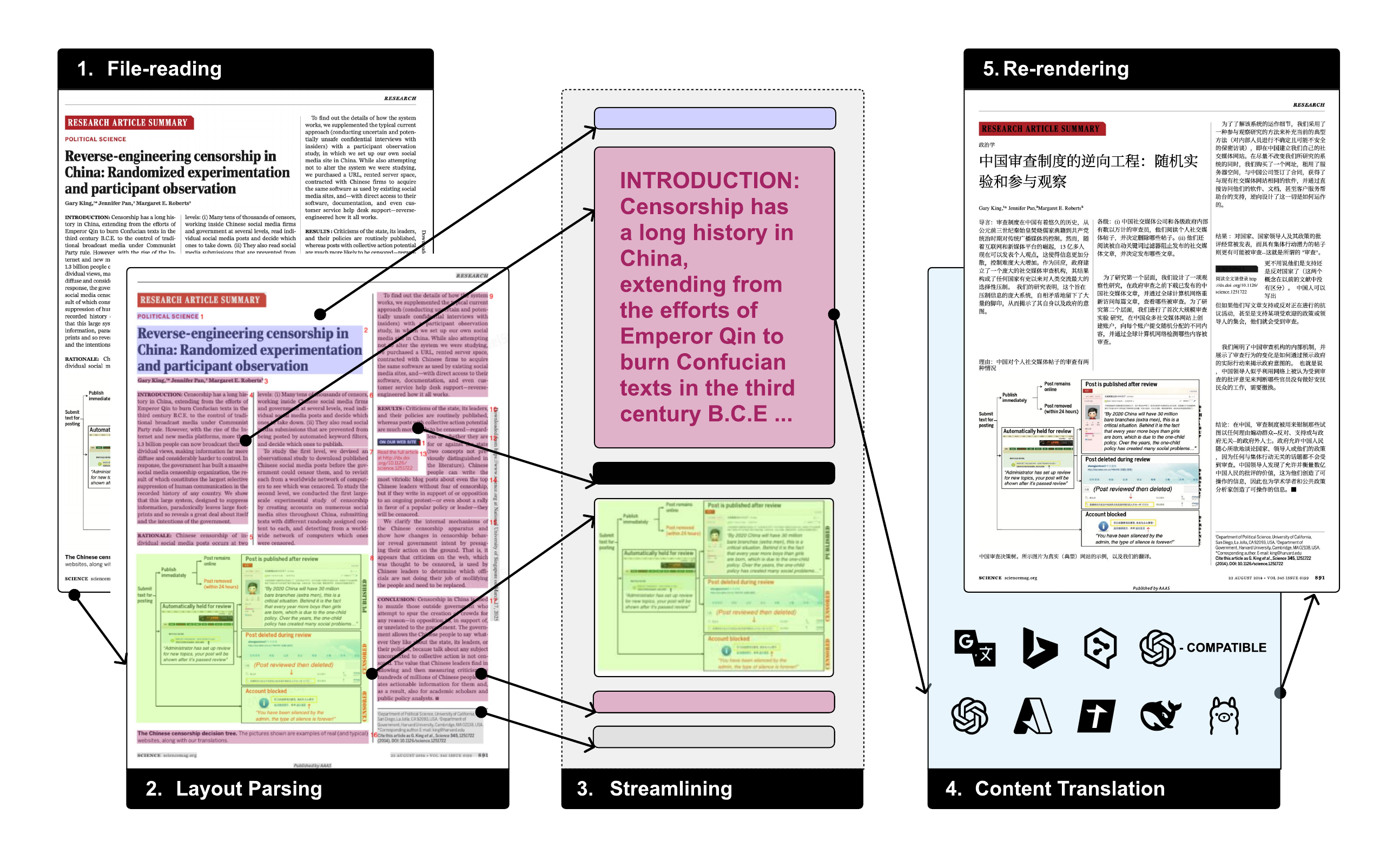}
    \caption{The architecture of PDFMathTranslate}    \label{fig:arch}
\end{figure*}

To fill the gap, we introduce \textbf{PDFMathTranslate}, the world's first open-source tool to translate PDF documents with preserved layouts. By leveraging recent advances in layout detection and large language models (see~\hyperref[fig:arch]{Figure 1}), we better address the barrier with at least five key contributions: (1) efficient workflow of layout detection, translation, and re-rendering; (2) support for multiple languages; (3) support for multiple translation models and services; (4) diverse user interfaces; and (5) a community-commerce model affording sustainable developments.

\section{Architecture and Design}\label{sec:arc}
PDFMathTranslate is designed to translate documents while preserving their original layouts. First, it accepts a PDF document along with user-specified parameters, such as languages and the preferred translation service. Next, it detects the layouts to extract the layouts and textual contents from the document. Third, the texts are translated using the selected translation service, such as GPT-4, DeepL, Google, or Ollama (see~\hyperref[tab:lang_serv]{Appendix A}). Finally, the translated text and previously detected layouts are re-rendered as a translated document with preserved layouts.

Technically, we design the architecture for \textit{precision}, \textit{flexibility}, and \textit{efficiency}. These three principles are realized by a precise parser, a flexible translation middle-ware, and an efficient workflow, detailed in following subsections.

\subsection{Precise layout parser}\label{sec:arch:layout}
Preserving layouts requires precise layout extraction. To precisely parse the position of document elements, we propose a pipeline consisting of layout detection, splitting, processing, and re-rendering. In the beginning, we exploit a recent advance in layout detection, ~\textit{DocLayout-YOLO-DocStructBench-onnx}. The model is another version of a SOTA solution in object detection, \textit{Yolov10}~\cite{wang2024yolov10}, and is fast and accurate in the specific task~\cite{zhao2024doclayoutyolo}.

To increase the compatibility of this model, we support two model formats: the \textit{Open Neural Network Exchange (ONNX)} \footnote{See \url{https://onnx.ai/}} standard and a \textit{pytorch} version. The ONNX version was chosen as default to ensure the compatibility of our parsing pipeline for diverse hardware.

\subsection{Flexible translation middleware}\label{sec:arch:service}
Translation middleware offers flexibility in terms of \textit{language diversity}, \textit{supported services}, and \textit{customization capabilities}. One, PDFMathTranslate supports at least 194 languages (see~\hyperref[tab:lang_serv]{Appendix A}), including popular ones such as \textit{English}, and some languages shared by relatively smaller communities, such as \textit{Cherokee}. Users can freely select any of these languages as either the input or output language when translating documents.

Two, PDFMathTranslate supports at least 23 popular services for translation. The list of services includes prominent online services such as Google Translate and OpenAI, as well as local deployments of models for either translation or dialogs (see~\hyperref[tab:lang_serv]{Appendix A}). In addition to explicitly listed services, users can integrate any new service complying with OpenAI protocols~\footnote{See: \href{https://platform.openai.com/docs/api-reference/chat/create}{https://platform.openai.com/docs/api-reference/chat/create}}.

Moreover, to simplify the addition of new services, we designed a structure that separates layout flows from translation flows. In this structure, language services receive only textual inputs, enhancing the generalizability and sustainability of our project. As a result, implementing any language service requires only a concise function of fewer than 15 lines.

Three, PDFMathTranslate provides additional customizability in translation. We designed a customization feature prompting strategies that potentially increase the quality of translation or its adaptability in domain-specific tasks. This feature allows users to exploit advanced prompting strategies such as few-shots~\cite{brown2020language}, Chain-of-Thought~\cite{wei2022chainofthought}, Role-playing~\cite{kong2024better, shanahan2023role}, etc. 

\subsection{Efficient streaming flow}\label{sec:arch:flow}Translating scientific and technical documents often necessitates an efficient workflow, especially when dealing with thousands of pages, which can challenge overall performance. To address the issue, we landed on a streaming design that is centered around an efficient in-memory processing pipeline that minimizes disk I/O while preserving document fidelity. Specifically, the \texttt{translate\_stream} function accepts a byte stream of a PDF file and transforms it into a mutable document representation. This approach allows the system to dynamically apply modifications, such as converting to PDF/A format when needed and embedding the appropriate fonts based on the target language. By employing temporary buffers and leveraging functions like \texttt{download\_remote\_fonts}, the design effectively manages resources and ensures that the original file remains unaltered during processing.

% 对于拓展性的说明
% 我们的工作流支持不同翻译后端，翻译器作为插件被设置在我们的工作流中。不同的翻译器通过继承基类并实现其中的do_translate方法，实现对不用翻译功能的接入。不同的翻译器可以快速的接入到不同的user interface中。目前在社区的贡献下，目前我们的工作流已支持20种以上的翻译器。同时对于openai兼容模式的后端LLM服务有着良好的拓展支持。
Our workflow supports various translation backends, with translators configured as plugins within our workflow. Different translators integrate distinct translation functionalities by inheriting from a base class and implementing the \texttt{do\_translate} method. These translators can be seamlessly integrated into different user interfaces. With contributions from the community, our workflow currently supports more than 20 translators. Additionally, it provides robust extensibility for backend LLM services with the OpenAI-compatible API and traditional machine translation service.

Further optimization in performance is achieved through a combination of asynchronous execution, caching strategies, and error-handling mechanisms. Asynchronous constructs, such as \texttt{asyncio.Event}, are employed to support task cancellation and concurrency, particularly when processing large documents or handling multiple translation tasks simultaneously. In addition, the design incorporates a caching mechanism—controlled via the \texttt{ignore\_cache} flag—to prevent redundant computations, save LLM tokens, and accelerate the translation process. While a subset font embedding strategy minimizes the final file size without compromising compatibility. Overall, this streaming design reflects a robust, scalable, and resource-efficient framework tailored for automated PDF translation workflows.

\section{Usage and Deployments}\label{sec:use}
PDFMathTranslate has been implemented in various interfaces, including a command-line tool (CLI), a graphic user interface (GUI), cross-platform applications (on MacOS and Windows), and Docker images. In addition to those officially supported interfaces, user \texttt{guaguastandup} contributes a Zotero plugin to our community.

\begin{figure}
  \centering
  \includegraphics[width=1\linewidth]{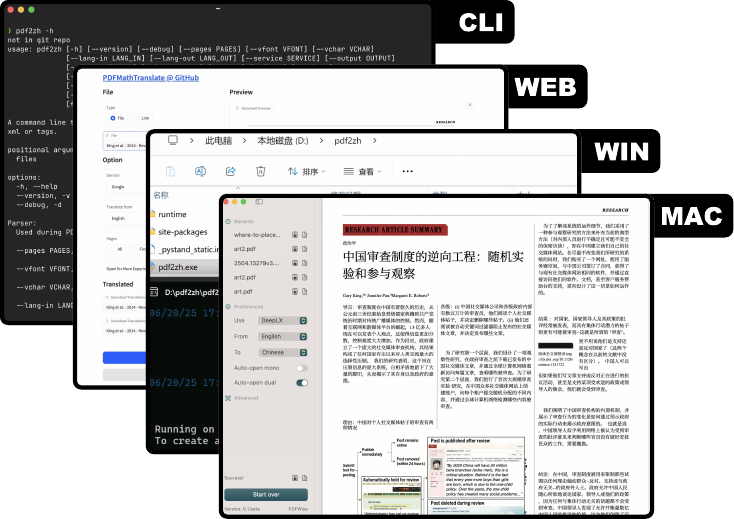}
  \caption{Officially supported interfaces}
  \label{fig:gui}
\end{figure}

For laymen, we offer GUI on Mac~\footnote{See~\href{https://github.com/reycn/pdf2zh-mac}{https://github.com/reycn/pdf2zh-mac}.}, Windows, Web, online demos, and Zotero plugin~\footnote{See~\href{https://github.com/guaguastandup/zotero-pdf2zh}{https://github.com/guaguastandup/zotero-pdf2zh}; community-contributed.}. Specifically, desktop versions on Mac and Windows can be installed, demos are publicly accessible online, and GUI can be started using \texttt{pdf2zh -i}. Those graphic interfaces are shown in Figure~\ref{fig:gui}.

% todo: windows
\begin{figure}
    \centering
    \includegraphics[width=1\linewidth]{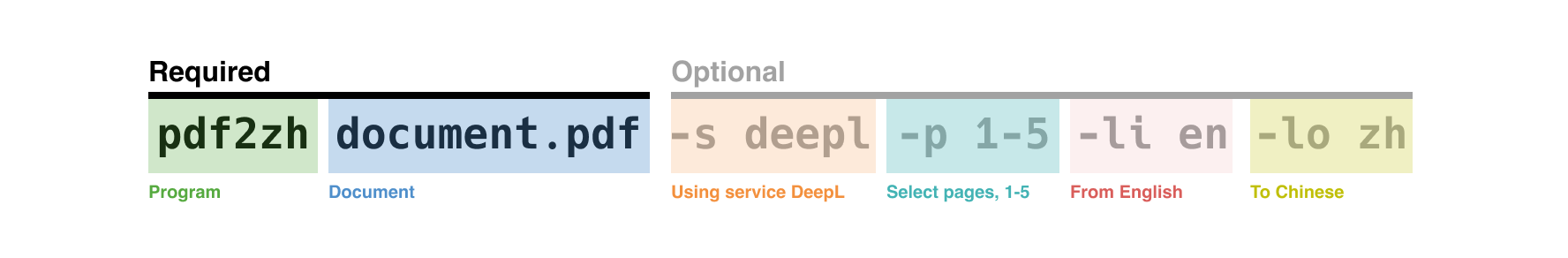}
    \caption{Major commands for CLI}
    \label{fig:cli}
\end{figure}

For developers, we support CLI and Docker for advanced usage and substream developments. Developers who prefer command-line tools can conveniently install our Python CLI program using \texttt{pip install pdf2zh}. The command-line tool supports extensive features, with the major options shown in Figure~\ref{fig:cli} and more documented in details~\footnote{See~\href{https://github.com/Byaidu/PDFMathTranslate/blob/main/docs/ADVANCED.md}{https://github.com/Byaidu/PDFMathTranslate/blob\\/main/docs/ADVANCED.md}}. For users who want to deploy the software on servers, we offer a Docker image for simplified deployment. Additionally, Docker images are built and distributed on different platforms for developers with potential network issues.
\begin{lstlisting}{language=bash}
  docker pull byaidu/pdf2zh
  docker run -d -p 7860:7860 byaidu/pdf2zh
\end{lstlisting}

% TODO: add cross ref table
\begin{table*}[hbp]
    \centering
    \scalebox{1}{
    \begin{threeparttable}
% \begin{tabular}{llrrrrrrr}
\begin{tabular}{llccccccc}
    \hline
    &                            & \textbf{OURS} & IMT\tnote{5} & Doc2X & TeX based\tnote{6} & Google & DeepL \\
    \hline
    \multirow{4}{*}{Accessibility}
    &Open-source                          & \cmark    &    &       &    \cmark          &        &       \\
    &Deployment                          & \cmark     &   &       &    \cmark          &        &       \\
    &API integration                    & \cmark     &    & \cmark &                   & \cmark & \cmark \\
    &Price                         & \textbf{Free}      & Paid     & Paid  & Free               & Free   & Paid  \\
    \hline
    \multirow{4}{*}{Readability}
    &Layout\tnote{1}           & \cmark  & \cmark        & \cmark&            &        &       \\
    &Formula\tnote{2}           & \cmark  & \cmark        & \cmark&    \cmark          &        &       \\
    &Bilingual\tnote{3}        & \cmark       & Partial   & Partial&   \cmark          & \cmark & \cmark\\
    &OCR\tnote{4}              & \cmark\tnote{7} & \cmark   & \cmark &                    &  \cmark  & \cmark \\
    \hline
    \multirow{2}{*}{Efficiency}
    &Batch tasks                       & \cmark    &      & \cmark   &  \cmark          & \cmark   & \cmark  \\
    &Speed (sec/page)           & \textbf{1.47}       & 1.50       & 1.86  & 1.67               & 0.38   & 1.88 \\
    \hline
\end{tabular}
    \begin{tablenotes}
      \footnotesize 
      \item[1] Whether the tool can preserve figures, images, and formulas (although the formulas may be incorrectly displayed).
      \item[2] Whether the tool can translate documents containing formulas and correctly handle the positions of the formulas.
      \item[3] Whether the tool supports bilingual exports where both the original and translated text are simultaneously readable.
      \item[4] Whether the tool can handle scanned documents instead of digital documents.
      \item[5] IMT: Immersive Translate PDF Pro.
      \item[6] The original LaTeX file is required for translation.
      \item[7] Available in a community fork: \texttt{PDFMathTranslate/PDFMathTranslate-next}. 
    \end{tablenotes}
    \end{threeparttable}}
    \caption{Comparison with other projects and productions}
    \label{tab:Comparison}
\end{table*}

% Our comparison indicates that PDFMathTranslate surpasses alternative solutions. First, regarding \textit{accessibility}, we significantly reduce user costs by providing an open-source, self-deployable, and free solution with flexible API integration. This flexibility enables broad applicability of our tool and ensures accurate translation outputs from any preferred service or model. Second, in terms of \textit{readability}, our approach preserves basic layouts, accurately recognizes complex mathematical formulas, and supports bilingual outputs. Finally, although our solution is not the fastest—due to the computational demands of on-device inference—its \textit{efficiency} remains comparable to other services (see \ref{tab:Comparison}). These advantages suggest that our tool is a better alternative to existing tools for translating scientific and technical documents, thereby potentially reducing language barriers more effectively.

\section{Comparison with Existing Systems}\label{sec:compare}
% PDFMathTranslate was designed to improve alternative commercial and open-source solutions. The ultimate goal of of this software is to better translate the scientific documents with precisely preserved layouts.

Compared with existing alternatives, PDFMathTranslate offers a superior solution from three perspectives: \textit{accessibility}, \textit{readability}, and \textit{efficiency} (see Table~\ref{tab:Comparison}). 

In terms of \textit{accessibility}, the tool is open-sourced, self-deployable, supports API calls, and is entirely free of charge. These advantages ensure wider accessibility compared to any other existing solutions. Regarding \textit{readability}, the tool preserves layouts, supports complex formulas, and is capable of exporting bilingual documents. Notably, a current limitation of our tool is its lack of optimal optical character recognition (OCR) support, which restricts its wider application for scanned documents. However, OCR functionality is planned for implementation in upcoming versions. Finally, concerning \textit{efficiency}, while this application is slower than text-based translation services (like Google Translate) due to the integration of precise layout detection models, our solution is significantly faster than alternative products that also preserve layouts (such as IMT and Doc2X).

\section{Use Cases}
\begin{figure*}[!hbp]
    \centering
    \includegraphics[width=0.9\linewidth]{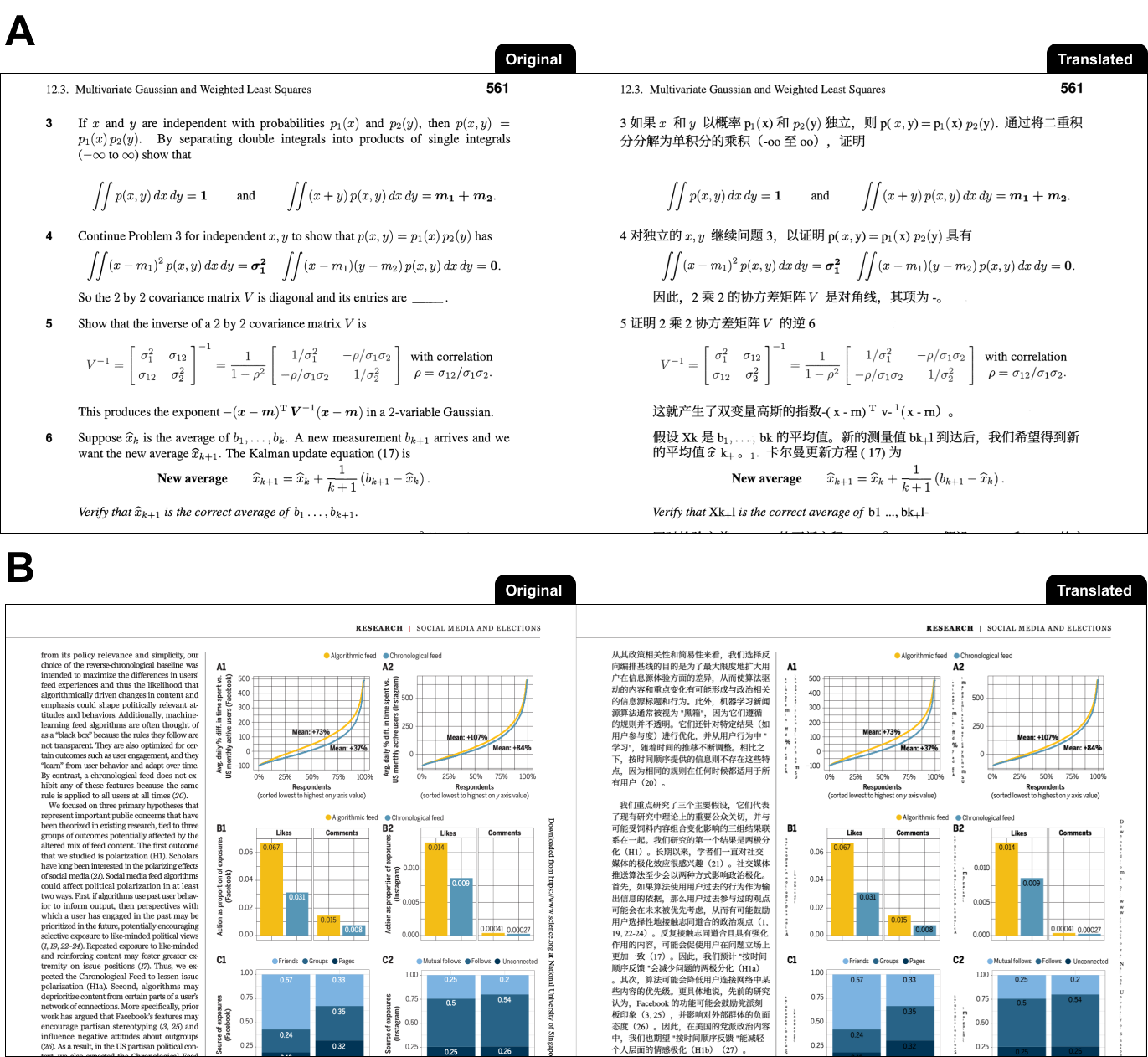}
    \caption{Use cases.}
    \label{fig:use-case}
\end{figure*}

We illustrate two typical use cases of PDFMathTranslate to illustrate how the tool translates text while preserving information embedded in layouts. The first use case (Panel A, Figure~\ref{fig:use-case}) is an excerpt from a textbook with both text and complex formulas, and the second case (Panel B, Figure~\ref{fig:use-case}) is a scientific research article with complex layouts and figures (see Figure~\ref{fig:use-case}). Both cases consistently show that our work is capable of precisely translating the information within text while preserving the crucial information embedded in layouts.

\section{Sustainability}
Nevertheless, maintenance is challenging to open-source projects~\cite{stol2010challenges}. To ensure sustainable development, we've established a \textit{community-commerce} model, through which the incentives of developers are provided by two sources: open-source recognition and the benefits of commercial products.

First, we increase developer exposure by prominently featuring contributors on the project's homepage and regularly highlighting recent contributions (e.g., \textit{1970-01-01 / Supports Google translation (by @author\_handle}). This strategy has successfully incentivized 44 global developers who collectively contributed tens of thousands of lines of code.

Second, we provide sponsored rewards such as membership exchanges for active contributors through collaboration with a commercial partner. These 11 subscriptions sent out to developers have effectively encouraged consistent contributions, resulting in the resolution of over 485 user-reported issues (by April 2025). Such cooperation has enabled the integration of advanced academic and community-driven technologies, including ONNX model support~\cite{Wybxc2025} and Qwen~\cite{ws0516822025}, thus advancing cross-linguistic scientific communication.

This sustainable collaboration model propelled the project to the top of GitHub's global trends for over a week, garnering more than 25k stars, 222k downloads, and over 49k Docker pulls (by June 2025). Additionally, our project's success has inspired a range of new commercial products with similar functionalities, significantly impacting both the scientific community and the broader public.

\section{Limitations}
While we have made several key improvements, our tool still has specific limitations. First, translation quality remains highly dependent on the underlying translation models and prompts. Second, the accuracy of the layout detection model in identifying various document layouts similarly depends on the quality of the layout detection models employed. Third, the current version of the tool can not handle scanned PDFs lacking optimal optical character recognition (OCR); however, this last feature is planned for implementation.

\section{Acknowledgments}
The authors thank all 44 contributors\footnote{They are \texttt{7shi, Byaidu, Copilot, Cycloctane, Hanaasagi, IuvenisSapiens, JEFF-dev-ui, Tql-ws1, Wybxc, YadominJinta, Zxis233, alohays, aseaday, awwaawwa, borcation, charles7668, chiu0602, czz404, damaoooo, dependabot[bot], domonnss, eltociear, hellofinch, highkay, hotwa, imClumsyPanda, kharkover, kidach1, lintian233, mydreamworldpolly, namazuchin, qqueing, reycn, tastelikefeet, timelic, treeleaves30760, tylzh97, ws051682, wx-11, xxnuo, xyzyx233, yidasanqian, ymattw, zqqian}; ordered alphabetically, retrieved through the official GitHub API by March 2025)} of the open-sourced project on GitHub; platforms allowing us to host demos (Hugging Face, ModelScope); services we used in the demo (e.g., Google Translate); redeem codes from Immersive Translate; public API access from SiliconFlow; and all comments from users.

\section{Ethics and Broader Impacts}
Our work has made a significant positive impact within both multilingual scientific and open-source communities. However, as a flexible, widely used tool, it raises potential ethical concerns regarding the copyright of documents. Translating documents without proper permissions could challenge the intellectual property rights of scientific works and innovations. We welcome suggestions from experts in intellectual property to mitigate the concern.

\bibliography{references}
\appendix
\section{Appendix}\label{sec:appendix}
\subsection{Officially supported language and services}\label{appendix:new_service}
In Table~\ref{tab:lang_serv}, we enumerate the officially supported services, models, and languages. However, the actual number may be significantly higher due to the extensibility of services and models.
\begin{table}[hbp!]
    \centering
    \caption{Supported languages and services for translation in PDFMathTranslate}
    \label{tab:lang_serv}
    \begin{tabular}{lll}
    \hline
        Category& Details & Total \\
        \hline
        Service or & Google \textit{(default)}, 302.AI, Bing, DeepL, DeepLX, Ollama,  Ali Qwen, & \textit{23} \tnote{1}  \\
        models & Ollama, Xinference, Gemma, OpenAI, OpenAI-like \tnote{1}, DeepSeek,\\ 
                & AzureOpenAI, Zhipu, ModelScope, Silicon Cloud, Gemini, Azure,\\
                & Tencent, Dify, AnythingLLM, Argos Translate, Grok, Groq\\
                % & \\
        \hline
        Input and & Abkhaz, Acehnese, Acholi, Afrikaans, Albanian, Alur, Amharic,  & \textit{194} \tnote{2}  \\
        output    &  Arabic, Armenian, Assamese, Awadhi, Aymara, Azerbaijani, Balinese, \\ 
        languages &  Bambara, Bashkir, Basque, Batak Karo, Batak Simalungun, \\
                  & Batak Toba, Belarusian, Bemba, Bengali, Betawi, Bhojpuri, \\
                  & Bikol, Bosnian, Breton, Bulgarian, Buryat, Cantonese, \\
                  & Catalan, Cebuano, Chichewa (Nyanja), Chinese (Simplified), \\
                  & Chinese (Traditional), Chuvash, Corsican, Crimean Tatar, \\
                  & Croatian, Czech, Danish, Dinka, Divehi, Dogri, Dombe, \\
                  & Dutch, Dzongkha, English, Esperanto, Estonian, Ewe, \\
                  & Fijian, Filipino (Tagalog), Finnish, French, \\
                  & French (French), French (Canadian), Frisian, \\
                  & Fulfulde, Ga, Galician, Ganda (Luganda), Georgian, \\
                  & German, Greek, Guarani, Gujarati, Haitian Creole, \\
                  & Hakha Chin, Hausa, Hawaiian, Hebrew, Hiligaynon, \\
                  & Hindi, Hmong, Hungarian, Hunsrik, Icelandic, Igbo, \\
                  & Iloko, Indonesian, Irish, Italian, Japanese, Javanese, \\
                  & Kannada, Kapampangan, Kazakh, Khmer, Kiga, Kinyarwanda, \\
                  & Kituba, Konkani, Korean, Krio, Kurdish (Kurmanji), \\
                  & Kurdish (Sorani), Kyrgyz, Lao, Latgalian, Latin, etc.\\
        \hline
    \end{tabular}

    \justifying
    \begin{tablenotes}
      \footnotesize
        \item[1] $^1$ The number of supported languages far exceeds what is shown in the table. Our protocol, designed as a general framework for OpenAI-like services, allows users to integrate any translation system, including locally hosted models or private online services, ensuring extensive language support.
        \item[2] $^2$ Based on the default service (Google Translate) and interface (CLI).
    \end{tablenotes}
\end{table}
\end{document}